# Heidelberg Colorectal Data Set for Surgical Data Science in the Sensor Operating Room

Lena Maier-Hein[1x], Martin Wagner[2x], Tobias Ross[1,3], Annika Reinke[1,3], Sebastian Bodenstedt[4], Peter M. Full[3,5], Hellena Hempe[1], Diana Mindroc-Filimon[1], Patrick Scholz[1,6], Thuy Nuong Tran[1], Pierangela Bruno[1,7], Anna Kisilenko[2], Benjamin Müller[2], Tornike Davitashvili[2], Manuela Capek[2], Minu Tizabi[1], Matthias Eisenmann[1], Tim J. Adler[1], Janek Gröhl[1], Melanie Schellenberg[1], Silvia Seidlitz[1,6], T. Y. Emmy Lai[5], Bünyamin Pekdemir[1], Veith Roethlingshoefer[8], Fabian Both[8,9], Sebastian Bittel[8,10], Marc Mengler[8], Lars Mündermann[11], Martin Apitz[2], Annette Kopp-Schneider[12], Stefanie Speidel[4,13], Hannes G. Kenngott[2x], Beat P. Müller-Stich[2x]

[x] Authors contributed equally

[1] Division of Computer Assisted Medical Interventions (CAMI), German Cancer Research Center (DKFZ), Im Neuenheimer Feld 223, 69120 Heidelberg, Germany
[2] Department for General, Visceral and Transplantation Surgery, Heidelberg University Hospital, Im Neuenheimer Feld 110, 69120 Heidelberg, Germany
[3] University of Heidelberg, Seminarstraße 2, 69117 Heidelberg, Germany
[4] Division of Translational Surgical Oncology, National Center for Tumor Diseases, Partner Site Dresden, Fetscherstraße 74, 01307 Dresden, Germany
[5] Division of Medical Image Computing (MIC), Im Neuenheimer Feld 223, 69120 Heidelberg, Germany
[6] HIDSS4Health – Helmholtz Information and Data Science School for Health, Im Neuenheimer Feld 223, 69120 Heidelberg, Germany
[7] Department of Mathematics and Computer Science, University of Calabria, Via Pietro Bucci, 87036 Arcavacata, Rende CS, Italy
[8] understandAI GmbH, An der RaumFabrik 34, 76227 Karlsruhe, Germany
[9] International Max Planck Research School for Intelligent Systems Tübingen, University of Tübingen, Geschwister-Scholl-Platz, 72074 Tübingen, Germany
[10] BMW Group, Heidemannstraße 164, 80939 Munich, Germany
[11] Corporate Research and Technology, Data-Assisted Solutions, KARL STORZ SE & Co. KG, Dr.-Karl-Storz-Strasse 34, 78532 Tuttlingen, Germany
[12] Division of Biostatistics, German Cancer Research Center (DKFZ), Im Neuenheimer Feld 581, 69120 Heidelberg, Germany
[13] Centre for Tactile Internet with Human-in-the-Loop (CeTI), TU Dresden, 01307 Dresden, Germany

Corresponding authors: Lena Maier-Hein (l.maier-hein@dkfz.de) and Beat P. Müller-Stich (beat.mueller@med.uni-heidelberg.de)

## Abstract
Image-based tracking of medical instruments is an integral part of surgical data science applications. Previous research has addressed the tasks of detecting, segmenting and tracking medical instruments based on laparoscopic video data. However, the proposed methods still tend to fail when applied to challenging images and do not generalize well to data they have not been trained on. This paper introduces the Heidelberg Colorectal (HeiCo) data set - the first publicly available data set enabling comprehensive benchmarking of medical instrument detection and segmentation algorithms with a specific emphasis on method robustness and generalization capabilities. Our data set comprises 30 laparoscopic videos and corresponding sensor data from medical devices in the operating room for three different types of laparoscopic surgery. Annotations include surgical phase labels for all video frames as well as information on instrument presence and corresponding instance-wise segmentation masks for surgical instruments (if any) in more than 10,000 individual frames. The data has successfully been used to organize international competitions within the Endoscopic Vision Challenges 2017 and 2019.

## Background & Summary

Surgical data science was recently defined as an interdisciplinary research field which aims "to improve the quality of interventional healthcare and its value through the capture, organization, analysis and modelling of data" [1]. The vision is to derive data science-based methodology to provide physicians with the right assistance at the right time. One active field of research consists in analyzing laparoscopic video data to provide context-aware intraoperative assistance to the surgical team during minimally-invasive surgery. Accurate tracking of surgical instruments is a fundamental prerequisite for many assistance tasks ranging from surgical navigation [2] to skill analysis [3] and complication prediction. While encouraging results for detecting, segmenting and tracking medical devices in relatively controlled settings have been achieved [4], the proposed methods still tend to fail when applied to challenging images (e.g. in the presence of blood, smoke or motion artifacts) and do not generalize well (e.g. to other interventions or hospitals) [5]. As of now, no large (with respect to the number of images), diverse (with respect to different procedures and levels of image quality), and extensively annotated data set (sensor data, surgical phase data, segmentations) has been made publicly available, which impedes the development of robust methodology.

This paper introduces a new annotated laparoscopic data set to address this bottleneck. This data set comprises 30 surgical procedures from three different types of surgery, namely from proctocolectomy (surgery to remove the entire colon and rectum), rectal resection (surgery to remove all or a part of the rectum), and sigmoid resection (surgery to remove the sigmoid colon). Annotations include surgical phase information and information on the status of medical devices for all frames as well as detailed segmentation maps for the surgical instruments in more than 10,000 frames (Fig. 1). As illustrated in Fig. 1 and 2, the data set is well-suited to both developing methods for instrument detection and binary or multi-instance segmentation. It features various levels of difficulty including motion artifacts, occlusion, inhomogeneous lighting, small or crossing instruments and smoke or blood in the field of view (see Fig. 3 for some challenging examples).

In this paper, we shall use the terminology for biomedical image analysis challenges that was introduced in a recent international guideline paper [6]. We define a biomedical image analysis challenge as an open competition on a specific scientific problem in the field of biomedical image analysis. A challenge may encompass multiple competitions related to multiple tasks, for which separate results and rankings (if any) are generated. The data set presented in this paper served as a basis for the Robust Medical Instrument Segmentation (ROBUST-MIS) challenge [7] organized as part of the Endoscopic Vision (EndoVis) challenge (https://endovis.grand-challenge.org/) at the International Conference on Medical Image Computing and Computer Assisted Interventions (MICCAI) 2019. ROBUST-MIS comprised three tasks, each requiring participating algorithms to annotate endoscopic image frames (Fig. 2). For the binary segmentation task, participants had to provide precise contours of instruments, represented by binary masks, with '1' indicating the presence of a surgical instrument in a given pixel and '0' representing the absence thereof. Analogously, for the multi-instance segmentation task, participants had to provide image masks with numbers '1', '2', etc. which represented different instances of medical instruments. In contrast, the multi-instance detection task merely required participants to detect and roughly locate instrument instances in video frames. The location could be represented by arbitrary forms, such as bounding boxes.

Information on the activity of medical devices and the surgical phase was also provided as context information for each frame in the 30 videos. This information was obtained from the annotations generated as part of the MICCAI EndoVis Surgical workflow analysis in the sensor operating room 2017 challenge (https://endovissub2017-workflow.grand-challenge.org/).

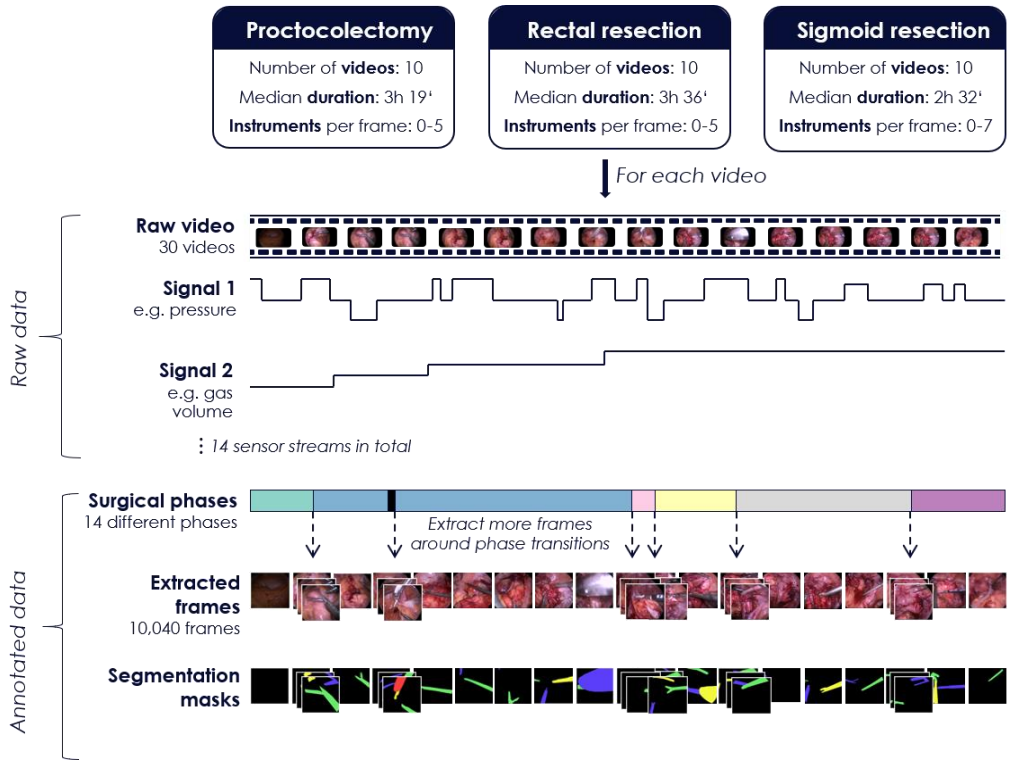

Fig. 1: Overview of the Heidelberg Colorectal (HeiCo) data set. Raw data comprises anonymized, downsampled laparoscopic video data from three different types of colorectal surgery along with corresponding streams from medical devices in the operating room. Annotations include surgical phase information for the entire video sequences as well as information on instrument presence and corresponding instance-wise segmentation masks of medical instruments (if any) for more than 10,000 frames.

**Methods**
The data set was generated using the following multi-stage process:
- I. Recording of surgical data
- II. Annotation of videos (surgical phases)
- III. Selection of frames for surgical instrument segmentation
- IV. Annotation of frames (surgical instruments)
  - A. Generation of protocol for instrument segmentation
  - B. Segmentation of instruments
  - C. Verification of annotations
- V. Generation of challenge data set

Details are provided in the following paragraphs.

I Recording of surgical data
Data acquisition took place during daily routine procedures in the integrated operating room KARL STORZ OR1 FUSION® (KARL STORZ SE & Co KG, Tuttlingen, Germany) at Heidelberg University Hospital, Department of Surgery, a certified center of excellence for minimally invasive surgery. Videos from 30 surgical procedures in three different types of surgery served as a basis for this data set: 10 proctocolectomy procedures, 10 rectal resection procedures, and 10 sigmoid resection procedures. While previous research on surgical skill and workflow analysis and corresponding publicly released data have focused on ex vivo training scenarios [8] and comparatively simple procedures, such as cholecystectomy [9], we placed emphasis on colorectal surgery. As these procedures are more complex, more variations occur in surgical strategy (e.g. length or order of phases) and phases may occur repeatedly.

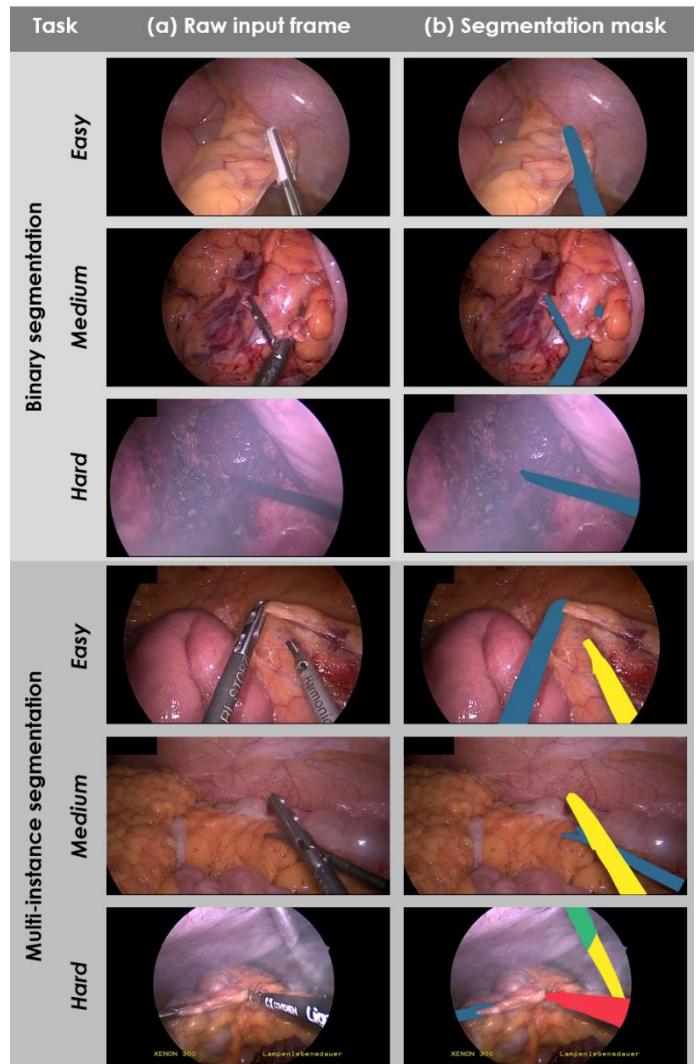

Fig. 2: Laparoscopic images representing various levels of difficulty for the tasks of medical instrument detection, binary segmentation and multi-instance segmentation. Raw input frames (a) and corresponding reference segmentation masks (b) computed from the reference contours.

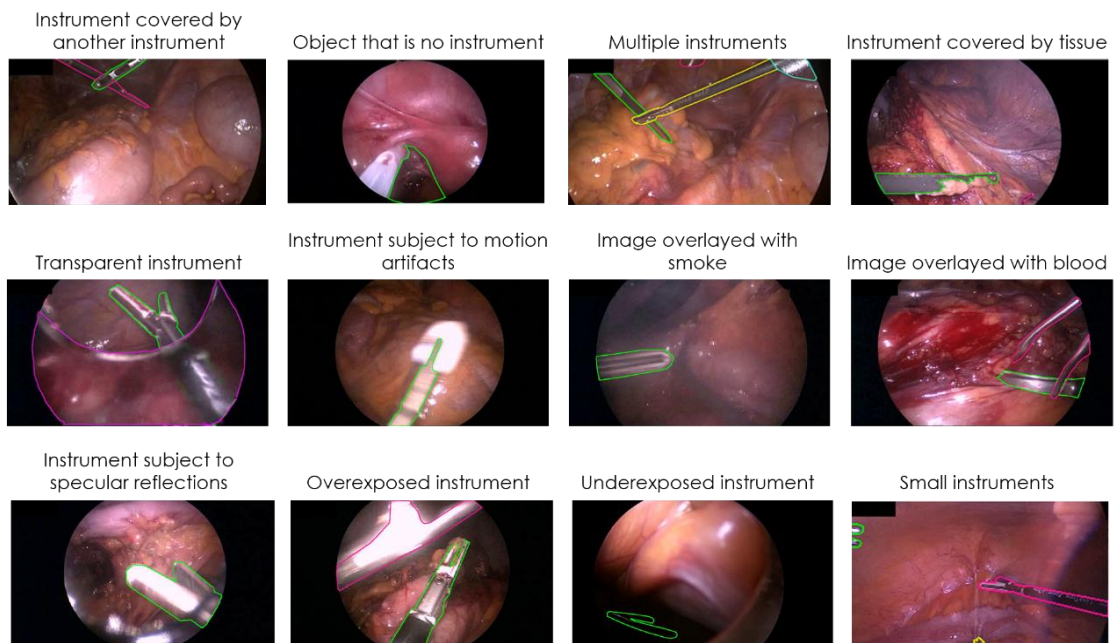

Fig. 3: Examples of challenging frames overlaid with reference multi-instance segmentations created by surgical data science experts.

All video data were recorded with a laparoscopic camera from KARL STORZ Image 1, with a forward-oblique telescope 30°. The KARL STORZ Xenon 300 was used as a cold light source. To comply with ethical standards and the general data protection regulation of the European Union, data were anonymized. To this end, frames corresponding to parts of the surgery performed outside of the abdomen were manually identified and subsequently replaced by blue images. Image resolution was scaled down from 1920x1080 pixels (high definition (HD)) in the primary video to 960x540 in our data set. In addition, KARL STORZ OR1 FUSION® was used to record additional data streams from medical devices in the room, namely Insufflator Thermoflator, OR lights, cold light fountain Xenon 300, and Camera Image 1. A complete list of all parameters and the corresponding descriptions can be found in Table 1.

It is worth noting that all three surgery types contained in this work included an extra-abdominal phase (bowel anastomosis; the connection of two parts of bowel) that was executed extra-abdominally without use of the laparoscope. As all three types of surgical procedure take place in the same anatomical region, many phases occur in two or all three of the procedures, as shown in Table 2.

II Annotation of videos
We use the following terminology throughout the remainder of this manuscript based on the definitions provided by [10]. Phases represent the highest level of hierarchy in surgical workflow analysis and consist of several steps. Steps are composed of surgical activities that aim to reach a specific goal. Activities represent the lowest level of hierarchy as "a physical task" or "well-defined surgical motion unit" such as dissecting, dividing or suturing.

In our data set, phases were modelled by surgical experts by first dividing the surgical procedure by dominant surgical activity, namely orientation (in the abdomen), mobilization (of colon), division (of vessels), (retroperitoneal) dissection (of rectum), and reconstruction with anastomosis. Subsequently, these parts were subdivided into phases by anatomical region. For example, the mobilization of colon was divided into phases for the sigmoid and descending colon, transverse colon, ascending colon and splenic flexure. Each phase received a unique ID, as shown in Table 2. Furthermore, during the annotation process, aberrations from the defined standard phase definitions occurred that had not been modelled beforehand. Examples include an additional cholecystectomy or a bladder injury. These phases were subsumed as "exceptional phases" (ID 13; see Table 2).

The annotator (surgical resident) had access to the endoscopic video sequence of the surgical procedure. The result of the annotation was a list of predefined phases for each video (represented by the IDs provided in Table 2) with corresponding timestamps denoting their starting points. The labeling was performed according to the following protocol:
1. Definition of the start of a phase
    a. Phases start when the instrument related to the first activity relevant for this phase enters the screen. Example: a grasper providing tissue tension for dissection of the sigmoid mesocolon in order to identify and dissect the inferior mesenteric artery.
    b. If a change of the anatomical region (such as change from mobilization of ascending colon to mobilization of transverse colon) results in the transition to a new phase, the camera movement towards the new region marks the start of the phase.
    c. If the camera leaves the body or is pulled back into the trocar between two phases, the new phase starts with the first frame that does not show the trocar in which the camera is located.
2. Definition of the end of a phase:
    a. Phases are defined by their starting point. The end of a phase thus occurs when the next phase starts. This implies that idle time is assigned to the preceding phase.

Note that while phases in other surgeries, such as cholecystectomy, follow a rigid process, this is not the case for more complex surgeries, such as the ones subject to this data set. In other words, each phase can occur multiple times. Moreover, colorectal surgery comes with possible technical variations between centers, surgeons and procedures. For example, in sigmoid resection, some surgeons may choose a tubular resection of the mesentery over a central dissection of vessels and lymph nodes en bloc for benign disease, which results in completely omitting the respective phase.

III Selection of frames

From the 30 surgical procedures described above, a total of 10,040 frames were extracted for instrument segmentation. In the first step, a video frame was extracted every 60 seconds. In this process, blue frames included due to video anonymization (see Recording of data) were ignored. This resulted in a total of 4,456 frames (corresponding to the extracted IDs) for annotation. To reach the goal of annotating more than 10,000 video frames in total, it was decided to place a particular focus on interesting snippets of the video. Surgical workflow analysis is currently a very active field of research. For an accurate segmentation of a video into surgical phases, it requires the detection of the transition from one surgical phase to the next. For this reason, frames corresponding to surgical phase transitions were obtained in seven of the 30 videos (three from rectal resection, two from the other two types of surgery). More specifically, frames within 25 seconds of the phase transition (before and after) were sampled every second (again, excluding blue frames). This led to a doubling of the number of annotated frames. Statistics of the number of frames selected for the different procedures are provided in Table 3.

IV Annotation of frames

An initial segmentation of the instruments in the selected frames was performed by the company understand.ai (https://understand.ai/). To this end, a U-Net style neural network architecture [11] was trained on a small manually labeled subset of the data set. This network was then used to label the rest of the data set in a semi-automated way; based on pixel-wise segmentation proposals, a manual refinement was performed, following previous data annotation policies [4]. Based on this initial segmentation, a comprehensive quality and consistency analysis was performed and a detailed annotation protocol was developed, which is provided in the Supplementary Methods. Based on this protocol, the initial annotations were refined/completed by an annotation team of four medical students and 14 engineers. In ambiguous or unclear cases, a team of two engineers and one medical student generated a consensus annotation. For quality control, two medical experts went through all of the segmentation masks and reported potential errors which were then corrected by members of the annotation team. Final agreement on each label was generated by a team comprising a medical expert and an engineer. Examples of annotated frames are provided in Fig. 2 and 3.

V Generation of challenge data set

Typically, a challenge has a training phase. At the beginning of the training phase, the challenge organizers release training cases with the relevant reference annotations [6]. These help the participating teams in developing their method (e.g. by training a machine learning algorithm). In the test phase, the participating teams either upload their algorithm, or they receive access to the test cases without the reference annotations and submit the results that their algorithm has achieved for the test cases to the challenge organizers. To enable comparative benchmarking to be executed for this challenge paradigm, our data set was split into a training set and a test set. Our data set was arranged such that it allows for validation of detection/binary segmentation/multi-instance segmentation algorithms in three test stages:

- **Stage 1:** The test data are taken from the procedures (patients) from which the training data were extracted.
- **Stage 2:** The test data are taken from the exact same type of surgery as the training data but from procedures (patients) that were not included in the training data.
- **Stage 3:** The test data are taken from a different but similar type of surgery (and different patients) compared to the training data.

Following this concept, the data set was split into training and test data as follows:

- The data from all 10 sigmoid resection surgery procedures were reserved for testing in stage 3. We picked sigmoid resection for stage 3 as it comprised the lowest number of annotated frames and we aimed to come as close as possible to the recommended 80%/20% split [12] of training and test data.
- Of the 20 remaining videos corresponding to proctocolectomy and rectal resection procedures, 80% were reserved for training and 20% (i.e. two procedures of each type) were reserved for testing in stage 2. More specifically, the two patients with the lowest number of annotated frames were taken as test data for stage 2 (for both rectal resection and proctocolectomy). Again, the reason for this choice was to increase the size of the training data set compared to the test set.
- For stage 1, every 10th annotated frame from the remaining 2*(10-2) = 16 procedures was used.

In summary, this elicited to a total of 10,040 frames, distributed as follows:
- Training data: 5,983 frames in total (2,943 frames from proctocolectomy surgery and 3,040 frames from rectal resection surgery)

- Test data (4,057 frames in total):
  - Stage 1: 663 frames in total (325 frames from proctocolectomy surgery and 338 frames from rectal resection surgery)
  - Stage 2: 514 frames in total (225 frames from proctocolectomy surgery and 289 frames from rectal resection surgery)
  - Stage 3: 2,880 frames from sigmoid resection surgery

As suggested in [6], we use the term case to refer to a data set for which the algorithm(s) participating in a specific challenge task produce one result (e.g. a segmentation map). To enable instrument detection/segmentation algorithms to take temporal context into account, we define a case as a 10 second video snippet comprising 250 endoscopic image frames (not annotated) and an annotation mask for the last frame (Fig. 4). In the mask, a '0' indicates the absence of a medical instrument and numbers '1', '2', ... represent different instances of medical instruments.

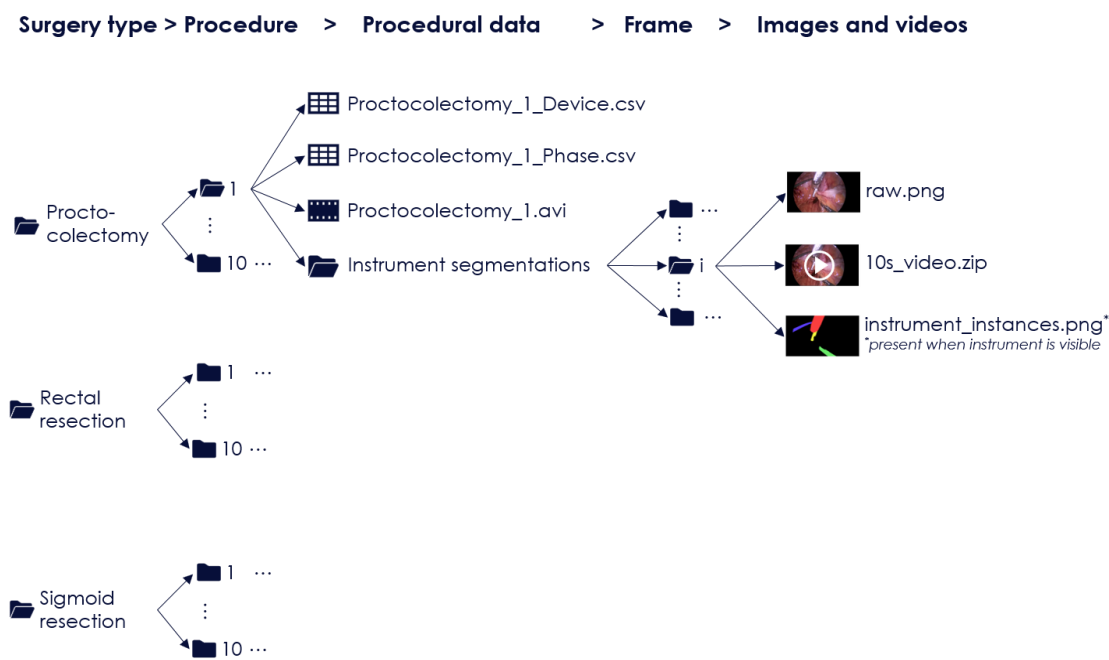

Fig. 4: Folder structure for the complete data set. It comprises five levels corresponding to (1) surgery type, (2) procedure number, (3) procedural data (video and device data along with phase annotations) and (4) frame number and (5) frame-based data.

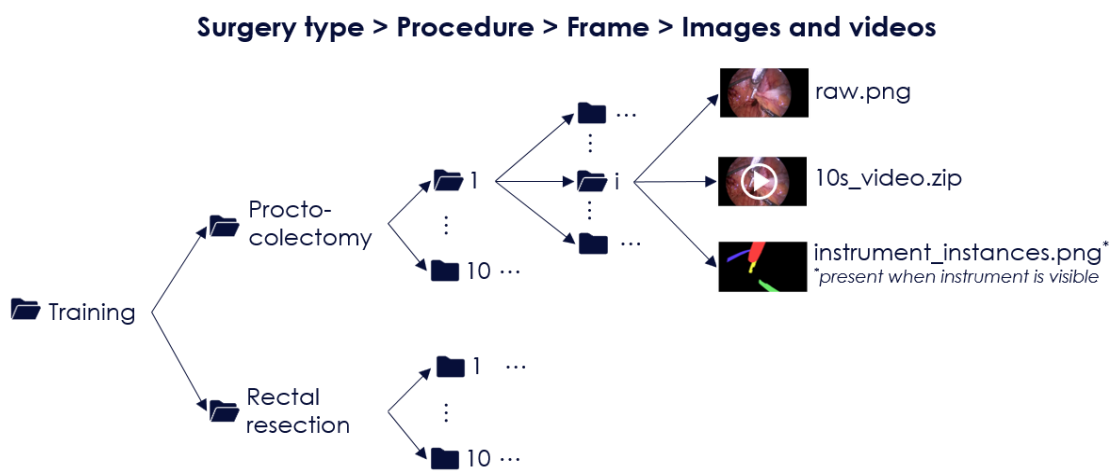

Fig. 5: Folder structure for the ROBUST-MIS challenge data set. It comprises five levels corresponding to (1) data type (training/test), (2) surgery type, (3) procedure number, (4) frame number and (5) case data.

**Data Records**
The primary data set of this paper corresponds to the ROBUST-MIS 2019 challenge and features 30 surgical videos along with frame-based instrument annotations for more than 10,000 frames. The annotations for this data set underwent a rigorous multi-stage quality control process. This data set is complemented by surgical phase annotations for the 30 videos which were used in the Surgical Workflow Analysis in the sensorOR challenge organized in 2017.

The data can be accessed in two primary ways: (1) As a complete data set that contains videos and medical device data along with corresponding annotations (surgical workflow and instrument segmentations) following the folder structure shown in Fig. 4 or (2) as ROBUST-MIS challenge data sets that represent the split of the data into training and test sets as used in the ROBUST-MIS challenge 2019 (Fig. 5).

I Complete data set
To access the complete data sets (without a split in training and test data), users are requested to create a Synapse account (https://www.synapse.org/). The data can be downloaded from [13].

The downloaded data is organized in a 4-level folder structure, as illustrated in Fig. 4. The first level represents the surgery type (proctocolectomy, rectal resection and sigmoid resection). In the next lower level, the folder names are integers ranging from 1-10 and represent procedure numbers. Each folder in this second level (corresponding to a surgery type p and procedure number i) contains the raw data (laparoscopic video as .avi file and device data as .csv file; see *Supplementary Table 1*), the surgical phase annotations (as .csv file, see *Supplementary Table 2*) and a set of subfolders numbered from 1 to Np,i where Np,i is the number of the frames for which instrument segmentations were acquired. The final 4th level represents individual video frames and contains the video frame itself (raw.png) and a 10 second video (10s_video.zip) of the 250 preceding frames in RGB format. If instruments are visible in an image frame, the folder contains an additional file called "instrument_instances.png", which represents the instrument segmentations generated according to the annotation protocol presented in the *Supplementary Methods*.

II ROBUST-MIS challenge data set
The segmentation data is additionally provided in the way it was available for participants of the ROBUST-MIS challenge [7]. Data usage requires the creation of a Synapse account. The data can be downloaded from [14]. Scripts to download the data and all scripts used for evaluation in the challenge can be found here: https://phabricator.mitk.org/source/rmis2019/.

The downloaded data is organized in a folder structure as shown in Fig. 5. There are two folders in the first level representing the suggested split into training and test data. The training data includes two folders representing the surgery type (proctocolectomy and rectal resection). The test data folder has three additional folders for stages 1-3. The next level for the test data also represents the surgery type (proctocolectomy, rectal resection, sigmoid resection). In the next deeper level of training and test data, the folder names are integers ranging from 1-10 and represent procedure numbers. Each folder in this level corresponds to a surgery type p and procedure number i and contains a set of Np,i subfolders where Np,i is the number of the frames for which instrument segmentations were acquired for the respective procedure and stage. The folders in the last level of the hierarchy contain the annotated video frame (raw.png) and a 10 second video (10s_video.zip) of the 250 preceding frames in RGB format. If instruments are visible in an image frame, the folder contains an additional file called "instrument_instances.png", which represents the instrument segmentations generated according to the annotation protocol presented in the Supplementary Methods. Raw data is stored in a separate folder and contains the videos (.avi file) as well as the device data (.csv file) for each procedure.

**Technical Validation**
I Validation of segmentations
The *verification* of the annotations was part of the data annotation procedure, as detailed above. To estimate the inter-rater reliability, five of the annotators that had curated the data set segmented (the same) 20 randomly selected images from the data set, where each image was drawn from a different surgery and contained up to three instrument instances. As we were interested in the inter-rater reliability for instrument instances rather than for whole images, we evaluated all 34 visible instrument instances of these 20 images individually. The Sørensen Dice Similarity Coefficient (DSC) [15] and the Haussdorf distance (HD) [16] were used as metric for contour agreement as they are the most widely used metrics for assessing segmentation quality [17]. For each instrument instance, we determined the DSC/HD for all combinations of two different raters. This yielded a median DSC of 0.96 (mean:

0.88, 25-quantile: 0.91, 75-quantile: 0.98) and a median HD of 12.8 (mean: 89.3, 25-quantile: 7.6, 25-quantile: 7.6, 75-quantile: 36.1) determined over all tuples of annotations and instrument instances. Manual analysis showed that outliers mainly occurred primarily if one or multiple of the raters did not detect a specific instance. It should be noted that an agreement of around 0.95 is extremely high given previous studies on inter-rater variability [18].

The recorded data and the corresponding segmentations/workflow annotations were used as basis for the ROBUST-MIS challenge 2019 [7] (https://phabricator.mitk.org/source/rmis2019/). According to the challenge results [7], the performance of algorithms decreases as the domain gap between training and test data increases. In fact, the performance dropped by 3% and 5% for the binary and multi-instance segmentation respectively (comparison of stage 1 with stage 3). This confirms our initial hypothesis that splitting the data set as suggested is useful for developing and validating algorithms with a specific focus on their generalization capabilities.

II Validation of surgical phase annotations

The phase annotations primarily serve as context information, which is why we did not put a focus on their validation. However, the following study was conducted to approximate intra-rater and inter-rater agreement for phases.

To assess the quality of the phase annotations, we randomly selected 10 time points in each of the 30 procedures resulting in n = 300 video frames. Then, we extracted a video snippet comprising 30 seconds before and 30 seconds after the respective frame from the video. The frames were independently categorized by five expert surgeons with at least 6 years of surgical experience, including the surgeon who performed the phase definition for our dataset in the first place, into the corresponding surgical phases according to our definition in Table 2. If the video snippet did not provide enough context to determine the phase, the surgeons reviewed the whole video.

For the statistical analysis, we compared the original annotation to the new annotation of the original rater (intra-rater agreement) and to the new annotation of the other surgeons (inter-rater agreement). Agreement between ground truth and raters was calculated as Cohen's kappa which quantifies agreement between two raters adjusted for agreement expected by chance alone. Calculation of unweighted kappa (for nominal ratings) with a 95% confidence interval (CI) was made by SAS Version 9.4 (SAS Inc., Cary, North Carolina, USA). To assess agreement between all five raters Fleiss' kappa for nominal ratings was performed with the function confIntKappa from the R package biostatUZH with 1,000 bootstraps (R Version 4.0.2, https://www.R-project.org). Values of kappa between 0.81 and 1.00 can be considered almost perfect. Intra-rater agreement between ground truth and new annotation by the original rater was 0.834 (CI 0.789 – 0.879) and inter-rater agreement between reference and each of the four other raters ranged from 0.682 (CI 0.626 – 0.739) to 0.793 (0.744 – 0.842). Accordingly, inter-rater agreement between reference and raters was at least substantial. Intra-rater agreement was almost perfect. Fleiss' kappa for agreement of all 5 raters was 0.712 (bootstrap 95% CI 0.673 - 0.747).

The recorded data and the corresponding segmentations/workflow annotations were used as the basis for the Surgical Workflow Analysis in the sensorOR 2017 challenge (https://endovissub2017-workflow.grand-challenge.org/).

The following data set characteristics have been computed based on the video and frame annotations. Eight of the 13 phases occurred in all three surgical procedures. The median (min;max) number of surgical phase transitions for proctocolectomy, rectal resection and sigmoid resection was 19 (15;25), 20 (10;31) and 14 (9;30) respectively. The median duration of the phases is summarized in Table 4. As shown in Table 5, the number of instruments per frame ranges from 0-7, thus reflecting the wide range of scenarios that can occur in clinical practice. Most frames (> 70% for all three procedures) contain only one or two instruments.

III Limitations of the data set

A limitation of our data set could be seen in the fact that the phase annotations were performed by only a single expert surgeon. It should be noted, however, that the phase annotations merely served as context information while the segmentations, which were generated with a highly quality-controlled process, are in the focus of this work. Furthermore, we acquired data from only one hospital. This implies limited variability with respect to the acquisition conditions as only one specific endoscope and light source were used. Still, to our knowledge, the HeiCo data set is the only publicly available data set (1) based on multiple different surgeries and (2) comprising not only annotated video data but also sensor data from medical devices in the operating room.

## Usage Notes

The data set was published under a Creative Commons Attribution-NonCommercial-ShareAlike (CC BY-NC-SA) license, which means that it will be publicly available for non-commercial usage. Should you wish to use or refer to this data set, you must cite this paper. The licensing of new creations must use the exact same terms as in the current version of the data set.

For benchmarking instrument segmentation algorithms, we recommend using the scripts provided for the ROBUST-MIS challenge (https://phabricator.mitk.org/source/rmis2019/) . They include Python files for downloading the data from the Synapse platform and evaluation scripts for the performance measures used in the challenge. For benchmarking surgical workflow analysis algorithms, we recommend using the script provided on Synapse [19] for the surgical workflow challenges.

To visualize the performance of an algorithm compared to state-of-the-art/baseline methods and/or algorithm variants, we recommend using the challengeR package [20] (https://github.com/wiesenfa/challengeR) written in R. It was used to produce the rankings and statistical analyses for the ROBUST-MIS challenge [7] and requires citation of the paper [20] for usage.

## Code Availability

The data set can be used without any further code. As stated in the usage notes, we recommend using the scripts provided for the ROBUST-MIS and surgical workflow challenges (https://phabricator.mitk.org/source/rmis2019/ and [19]) as well as the challengeR package [20] (https://github.com/wiesenfa/challengeR) for comparative benchmarking of algorithms.

## Acknowledgements

This project has been funded by the Surgical Oncology Program of the National Center for Tumor Diseases (NCT) Heidelberg and the project "OP4.1," funded by the German Federal Ministry of Economic Affairs and Energy (grant number BMWI 01MT17001C). In addition, the project has been funded by "InnOPlan", funded by the German Federal Ministry of Economic Affairs and Energy (grant number BMWI 01MD15002E). It was further supported by the Helmholtz Association under the joint research school HIDSS4Health (Helmholtz Information and Data Science School for Health) and the German Research Foundation (DFG) as part of Germany's Excellence Strategy - EXC2050/1 - Project ID 390696704 - Cluster of Excellence "Centre for Tactile Internet with Human-in-the-Loop" (CeTI).

## Author contributions

L.M.-H. initiated and coordinated the work on the segmentation data set, developed the concept for the instrument annotations, analyzed the data, co-organized the ROBUST-MIS challenge and wrote the labeling protocol (Supplementary Methods) as well as the manuscript.
M.W. acquired the data, developed the concept for the phase annotations, performed the phase annotations, co-organized the surgical workflow analysis in the sensorOR challenge and the ROBUST-MIS challenge, verified the instrument segmentations as a medical expert, performed the phase validation and wrote the manuscript.
T.R. and A.R. coordinated the work on the segmentation data set, developed the concept for the instrument annotations, analyzed the data, organized the ROBUST-MIS challenge, wrote the labeling protocol and contributed to the writing of the manuscript.
P.M.F., H.H., D.M.F., P.S, T.N.T., P.B. annotated the data for the ROBUST-MIS challenge and helped writing the labeling protocol.
A.K., Be.M., T.D., M.C., J.G., S.S., M.S., M.E., E.L., T.J.A., V.R., F.B., S.Bit. and M.M. annotated the data for the ROBUST-MIS challenge.
B.P. verified all annotations for the ROBUST-MIS challenge.
M.A. verified the instrument segmentations as a medical expert.
L.M. preprocessed, structured and analyzed the medical device data.
A.K.-S. coordinated and implemented the statistical analyses.
M.T. analyzed the data and contributed to the writing and proofreading of the manuscript.
S.Sp. and S.B. co-organized the surgical workflow analysis in the sensorOR challenge, developed the concept for the phase annotations, analyzed the surgical phase annotations and the medical device data, and contributed to the writing and proofreading of the manuscript.
F.N., H.G.K. and P.P. performed the phase validation.

H.G.K. and B.M. acquired the data, developed the concept for the phase annotations, (co-) organized the surgical workflow analysis in the sensorOR challenge and the ROBUST-MIS challenge, and proofread the manuscript.

**Competing interests**

L.M.-H., T.R., A.R., S.B. and S.Sp. worked with device manufacturer KARL STORZ SE & Co. KG in the joint research project "OP4.1," funded by the German Federal Ministry of Economic Affairs and Energy (grant number BMWI 01MT17001C). M.W., B.M. and H.G.K. worked with device manufacturer KARL STORZ SE & Co. KG in the joint research project "InnOPlan," funded by the German Federal Ministry of Economic Affairs and Energy (grant number BMWI 01MD15002E). F.N. reports receiving travel support for conference participation as well as equipment provided for laparoscopic surgery courses by KARL STORZ SE & Co. KG, Johnson & Johnson, Intuitive Surgical, Cambridge Medical Robotics and Medtronic. V.R, F.B, S.Bit. and M.M. are/were employees of the company understand.ai, which sponsored the initial labeling of the ROBUST-MIS data set. L.M. is employee of the company KARL STORZ SE & Co. KG, which sponsored a scientific prize for the surgical workflow analysis in the sensorOR challenge. The authors not listed as employees of understand.ai or KARL STORZ SE & Co. KG are not and have never been in conflict of interest or financial ties with the company. All other authors have no competing interests.

**Tables**

*Table 1: Medical devices of the operating room and corresponding sensor streams provided by KARL STORZ OR1 FUSION® (KARL STORZ SE & Co KG, Tuttlingen, Germany).*

| Medical device name | Medical device description | Sensor stream names |
| --- | --- | --- |
| Insufflator Thermoflator (KARL STORZ SE & Co. KG, Tuttlingen, Germany) | Device used to insufflate the abdomen with carbon dioxide in order to create space for the minimally invasive surgery. | Flow Actual |
| | | Flow Target |
| | | Pressure Actual |
| | | Pressure Target |
| | | Gas Volume |
| | | Supply Pressure |
| OR lights LED2 (Dr. Mach GmBH & Co KG, Germany) | Light mounted onto movable arms on the ceiling. Used to illuminate the patient's abdomen during open surgery. | Light1 On |
| | | Light1 Intensity Actual |
| | | Light2 On |
| | | Light2 Intensity Actual |
| Coldlight fountain Xenon 300 (KARL STORZ SE & Co KG, Tuttlingen, Germany) | Light source that illuminates the abdomen via a light cable mounted onto the laparoscopic camera. | Intensity Actual |
| | | Standby |

| | | | |
|---|---|---|---|
| Camera Image 1 (KARL STORZ SE & Co. KG, Tuttlingen, Germany) | Endoscopic camera control unit for use with both single and three-chip camera heads. | White Balance Shutter Speed Brightness Enhancement | |

*Table 2: Definition of surgical workflow phases and their occurrence in proctocolectomy (P), rectal resection (RR) and sigmoid resection (SR).*

| ID | Phase name | Phase definition | Present in |
|---|---|---|---|
| 0 | General preparation and orientation in the abdomen | This phase starts when the laparoscopic camera is first inserted into the abdomen. The phase may include removal of adhesions from the abdominal wall. The phase ends when the first instrument is inserted to manipulate the colon, which marks the beginning of a (colon) mobilization phase. | P, RR, SR |
| 1 | Dissection of lymph nodes and blood vessels en bloc | This phase starts when the mesocolon is dissected in order to identify and dissect the inferior mesenteric artery and vein. The phase may include a dissection of the mesocolon after the isolation and division of the vessels in order to harvest the corresponding lymph nodes en bloc. | P, RR, SR |
| 2 | Retroperitoneal preparation towards lower pancreatic border | This phase starts with the preparation of the retroperitoneal fat medially to the descending colon toward the lower pancreatic border. The aim of the preparation is a central dissection of the inferior mesenteric vein. | RR |
| 3 | Retroperitoneal preparation of duodenum and pancreatic head | This phase starts with the preparation of the retroperitoneal fat medially to the ascending colon towards the duodenum. This phase corresponds to phase 2, but on the site of the ascending colon. | P |

| 4 | Mobilization of sigmoid colon and descending colon | This phase starts with the manipulation of the sigmoid or descending colon (usually sigmoid). It comprises mobilization of both sigmoid and descending colon below the splenic flexure. The phase also includes positioning and a check for remaining adhesions or bleedings of the descending colon after mobilization. | P, RR, SR |
|---|---|---|---|
| 5 | Mobilization of splenic flexure | This phase usually follows phase 4 and starts as soon as the mobilization of the descending colon reaches the splenic flexure in the continuum of the bowel at the lower border of the spleen. During proctocolectomy it includes the retroperitoneal preparation towards the lower border of pancreas (phase 2 in rectal resection) | P, RR, SR |
| 6 | Mobilization of transverse colon | This phase starts with the mobilization of the transverse colon and includes the separation of the colon from the greater omentum. | P, RR, SR |
| 7 | Mobilization of ascending colon | This phase starts with the mobilization of the ascending colon, either after or before mobilizing the transverse colon. It includes the mobilization of the terminal ileum. | P |
| 8 | Dissection and resection of the rectum | This phase starts with the incision of the peritoneal fold towards the small pelvis or with an advancement toward the small pelvis after previous incision of the peritoneum for mobilization of the sigmoid or dissection of the vessels. This phase may include the isolation of the left ureter. For sigmoid resection it may include a circular preparation of the rectosigmoid for resection and stapling of the rectosigmoidal junction only. | P, RR, SR |
| 9 | Extra-abdominal preparation of anastomosis | This phase starts when the camera is retracted from the abdomen to perform the mini-laparotomy for extraction of the specimen. For proctocolectomy it includes the preparation of an ileal pouch. | P, RR, SR |

| 10 | Intra-abdominal preparation of anastomosis | This phase starts when the camera enters the abdomen after the extra-abdominal preparation of the anastomosis is finished. It includes irrigation and hemostasis (stopping of bleeding) in the small pelvis and other places in the abdomen after resection and after extra-abdominal preparation of the anastomosis. It may also include additional mobilisation of ileum, colon or (in sigmoid resection) remaining rectum to reduce tension on the anastomosis. | P, RR, SR |
|---|---|---|---|
| 11 | Creation of stoma | This phase starts when small bowel length is estimated beginning from the terminal ileum in order to identify the bowel loop for the stoma. The phase may include an extra-abdominal part for positioning of the stoma and afterwards a laparoscopic inspection for torquation of bowel. Also, the extra-abdominal suturing of the stoma may take place before final laparoscopic inspection or afterwards. | P, RR |
| 12 | Finalization of operation | This phase starts with and includes insertion of drains and / or irrigation and final check for bleedings. The phase may include inspection of small bowel for lesions and their suturing. | P, RR, SR |
| 13 | Exception | This phase comprises some unique phases that occurred exceptionally and unexpectedly during the annotation of the videos, due to the complexity of colorectal procedures. This included the closing of a bladder lesion, excisions of liver tissue for biopsy, appendectomy, a second resection of the rectum during the same procedure, resection of a mesocolonic cyst, cholecystectomy, elaborate hemostasis of splenic bleeding and a suturing of the abdominal wall due to bleeding from a trocar incision. | P, RR, SR |

*Table 3: Number of frames selected from the different procedures.*

| Surgery type | Number of videos | Number of annotated frames: median (min; max) | Number of frames with instruments: median (min; max) | Number of instruments per frame: median (min; max) |
|---|---|---|---|---|
| Proctocolectomy | 10 | 152 (101, 1259) | 133 (93, 1130) | 1 (0, 5) |
| Rectal resection | 10 | 198 (123, 1181) | 169 (107, 870) | 1 (0, 5) |
| Sigmoid resection | 10 | 121 (69, 1070) | 105 (62, 730) | 1 (0, 7) |

*Table 4: Median duration of phases [min]. IDs are introduced in Table 2.*

| | Phase ID | | | | | | | | | | | | | |
|---|---|---|---|---|---|---|---|---|---|---|---|---|---|---|
| **Surgery type** | **0** | **1** | **2** | **3** | **4*** | **5** | **6** | **7** | **8**** | **9** | **10** | **11** | **12** | **13** |
| Proctocolectomy | 7 | 7 | 0 | 4 | 13 | 4 | 17 | 6 | 51 | 42 | 23 | 10 | 4 | 0 |
| Rectal surgery | 7 | 19 | 12 | 0 | 30 | 6 | 3 | 0 | 46 | 33 | 15 | 5 | 8 | 2 |
| Sigmoid surgery | 6 | 11 | 0 | 0 | 34 | 0 | 2 | 0 | 13 | 34 | 16 | 0 | 7 | 0 |

*Phase 4 (mobilization of sigmoid colon and descending colon) is shorter for proctocolectomy because no oncological but tubular resection is performed.

** Phase 8 (dissection and resection of the rectum) is shorter for sigmoid resection because only the high part of the rectum, but not the middle and low part of the rectum are subject to removal.

*Table 5: Number of instruments in annotated frames. Most frames (> 70% for all three procedures) contain one or two instruments.*

| | Number of frames with *n* instruments | | | | | | | |
|---|---|---|---|---|---|---|---|---|
| **Surgery type** | ***n = 0*** | ***n = 1*** | ***n = 2*** | ***n = 3*** | ***n = 4*** | ***n = 5*** | ***n = 6*** | ***n = 7*** |
| Proctocolectomy | 450 (12.9%) | 1,697 (48.6%) | 1,063 (30.4%) | 227 (6.5%) | 54 (1.5%) | 2 (0.1%) | 0 (0.0%) | 0 (0.0%) |
| Rectal surgery | 714 (19.5%) | 1,850 (50.4%) | 917 (25.0%) | 158 (4.3%) | 21 (0.6%) | 7 (0.2%) | 0 (0.0%) | 0 (0.0%) |
| Sigmoid surgery | 650 (22.6%) | 1,198 (41.6%) | 827 (28.7%) | 178 (6.2%) | 24 (0.8%) | 2 (0.1%) | 0 (0.0%) | 1 (0.0%) |

# Supplementary Information Guide

## Supplementary Methods (p. 3-9)

The Supplementary Methods contain the labeling protocol for the Robust Medical Instrument Segmentation (ROBUST-MIS) Challenge 2019.

## Supplementary Tables (p. 10)

### Supplementary Table 1 (p. 10)

Supplementary Table 1 shows an example of the device data as it is provided in a .csv file for all 30 surgical procedures.

### Supplementary Table 2 (p. 11)

Supplementary Table 2 shows an example of the phase annotations as they are provided in a .csv file for all 30 surgical procedures.

# Supplementary Information

Heidelberg colorectal data set for surgical data science in the sensor operating room

Maier-Hein et al.

# Supplementary Methods

## ROBUST-MIS Labeling Instructions

### Introduction

Intraoperative tracking of laparoscopic instruments is often a prerequisite for computer and robotic assisted interventions. Although previous challenges have targeted the task of detecting, segmenting and tracking medical instruments based on endoscopic video images, key issues remain to be addressed:
- Robustness: The methods proposed still tend to fail when applied to challenging images (e.g. in the presence of blood, smoke or motion artifacts, different and new instruments).
- Generalization: Algorithms trained for a specific intervention in a specific hospital typically do not generalize.

The goal of this challenge is, therefore, the benchmarking of medical instrument detection and segmentation algorithms, with a specific emphasis on robustness and generalization capabilities of the methods. The challenge is based on the biggest annotated dataset made (to be made) publically available, comprising 10,000 annotated images that have been extracted from a total of 30 surgical procedures from three different surgery typesies.

### Terminology

**Matter:**
- Anything that has mass, takes up space and can be clearly identified.
- Examples: tissue, surgical tools, blood
- Counterexamples: reflections, digital overlays, movement artifacts, smoke

**Medical instrument to be detected and segmented:**
- Elongated rigid object put into the patient and manipulated directly from outside the patient
    - Examples: grasper, scalpel, (transparent) trocar, clip applicator, hooks, stapling device, suction
    - Counterexamples: non-rigid tubes, bandage, compress, needle (not directly manipulated from outside but manipulated with an instrument), coagulation sponges, metal clips

### Tasks

Participating teams may enter competitions related to the following tasks:

Binary segmentation:
- Input: 250 consecutive frames (10sec) of a laparoscopic video with the last frame containing at least one medical instrument
- Output: a binary image, in which "0" indicates the absence of a medical instrument and a number ">0" represents the presence of a medical instrument.

      Multiple instance detection and segmentation:
- 250 consecutive frames (10sec) of a laparoscopic video;
- Output: a binary image, in which "0" indicates the absence of a medical instrument and numbers "1", "2", ... represent different instances of medical instruments.

For both tasks, the entire corresponding video of the surgery is provided along with the training data as context information. In the test phase, only the test image along with the preceding 250 frames is provided.

## Labeling instructions

Annotators have access to the video frame to be annotated as well as to the video sequence (for both the training and the test phase!). Before making an annotation, they can scroll through the video sequence.

The results of the manual annotation are one or multiple closed contours as illustrated in the examples below. The interior of these contours represent a medical instrument. Everything outside the provided contours is regarded as other matter (not medical instrument) or image overlay. Contours of the same colour in one image represent all visible parts of one instance of an instrument.

The following decisions were made to ensure consistent labels:
1. Algorithm target: Only medical instruments as defined above should be segmented (see examples 4, 6, 10, 13).
2. Occlusion: Each pixel may correspond to exactly one structure. Specifically, the solid/liquid matter that occurs first along the line of sight of the endoscope determines the label. This may result in multiple contours for a single instrument that is occluded by another instrument, blood or tissue, for example. See examples 2 (instrument) and 9 (tissue) below.
3. Transparency: Medical instruments may be transparent. The occlusion rule holds in this case as well. See examples 8 and 14 below.
4. Holes in instruments: Several medical instruments feature holes (see example 3 below). A hole is made up of pixels that do not show parts of the instrument but are either a) completely surrounded by pixels of the same instrument or (b) are completely surrounded by pixels of one instrument and the margin of the image where it is known, from video context, that the instrument would close the hole outside the image. Following recommendations of previous challenges and given the difficulties of localizing these holes, they are regarded as part of the instrument (i.e. "inpainted"), as illustrated in example 3. The sole exception are trocars when the camera is placed inside of them (see example 14).
5. Text overlay: Text overlay (see example 17) shall be ignored.
6. Image overlay: Image overlays (see example 11 and 16) are treated like "other matter" (not part of instrument), i.e., they are regarded as the first object in the line of light. This also applies for the censor boxes in examples 2 and 5.

Tricky examples

| # | Image | How to |
|---|---|---|
| 1 | 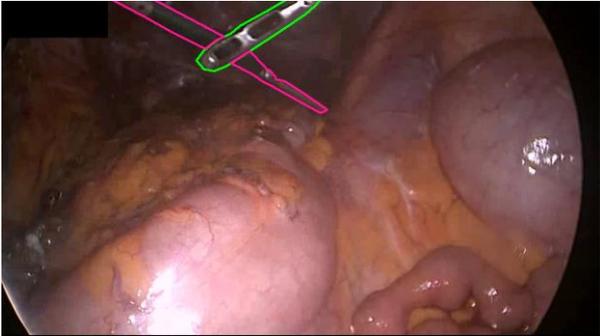 | Each pixel may correspond to exactly one structure; the solid/liquid matter that occurs first along the line of sight of the endoscope determines the label (rule 2). In this case, this results in two contours for the instrument with the pink contour.<br>Even if instruments (pink) are visible through holes of another instrument (green), they are not annotated as holes but are regarded as part of the (other) instrument (rule 4). |
| 2 | 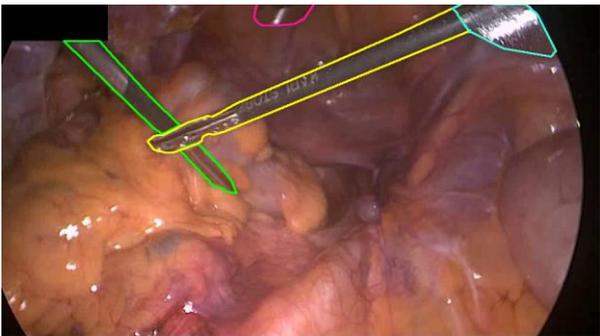 | This image contains four instances of medical instruments. The instrument with the green contour is partially occluded by an image overlay (black box - rule 6) and another instrument (yellow). The yellow one is partially occluded by a trocar (rule 2). |
| 3 | 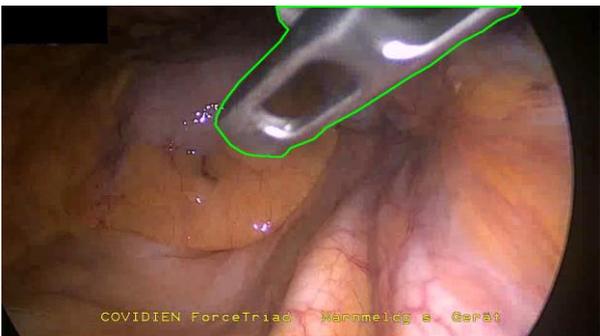 | The two holes (one barely visible in the corner) of the medical instrument are regarded as part of the instrument (rule 4). |
| 4 | 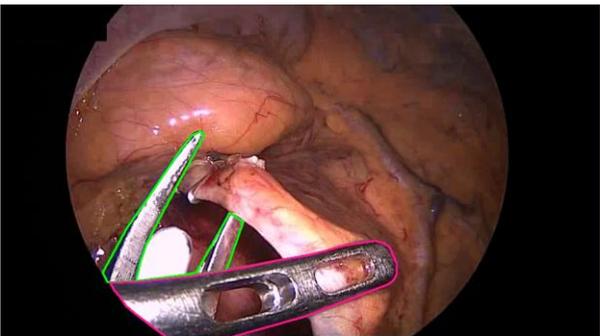 | This example does not only show another application of rule 1 and 4 but also emphasizes the distinction of medical instruments and other objects in rule 1 (the white plastic object is not part of an instrument). |
| 5 | 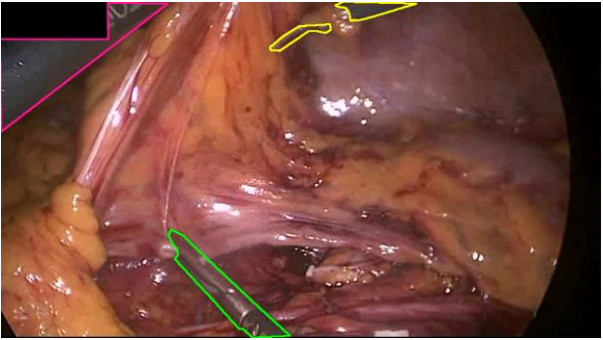 | Three instances of medical instruments, two of which are partially occluded (rules 2 and 6). |
| 6 | 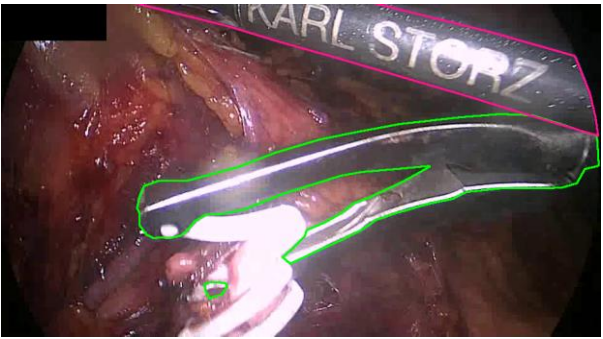 | Plastic clamps are not regarded as medical instruments and therefore not annotated (rule 1). |
| 7 | 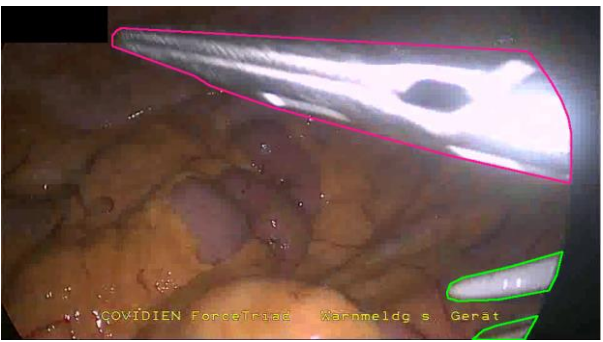 | Meta knowledge/video context information was used to conclude that the two object parts in the lower right corner correspond to one instrument. |
| 8 | 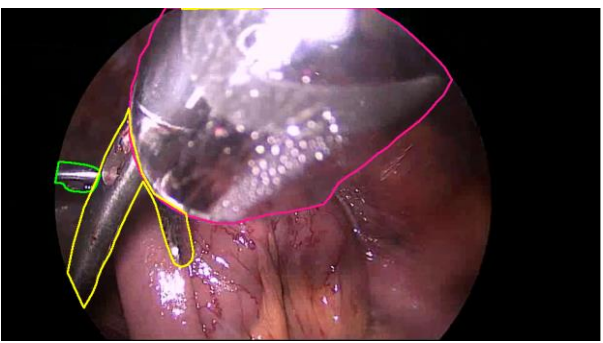 | Very challenging image. The transparent trocar occludes part of the instrument represented by the yellow contour (rule 3). |
| 9 | 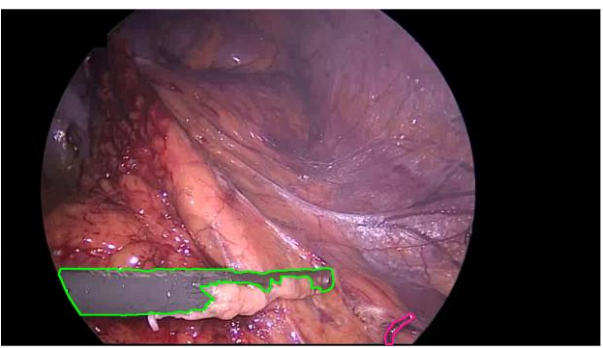 | Only visible parts of the instruments are annotated. The instrument corresponding to the green contour is occluded by tissue (rule 2). The one represented by pink is barely visible. |
| 10 | 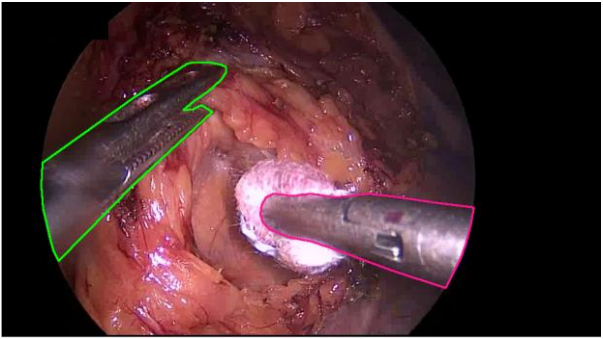 | Additional materials that are not regarded as medical instruments (here: bandage) are not segmented (rule 1). |
| 11 | 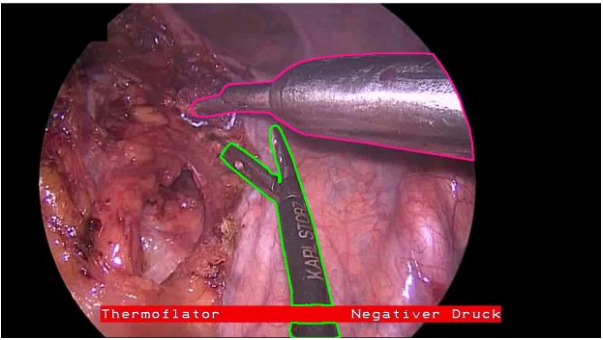 | The instrument represented by the green contour comprises two parts due to image overlay (rule 6). |
| 12 | 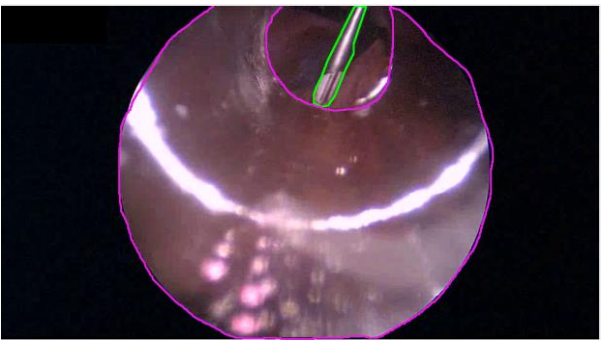 | Tricky example. A medical instrument is visible at the end (opening) of the trocar. |
| 13 | 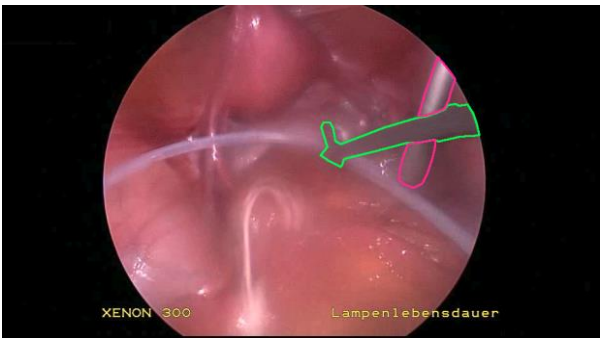 | Additional materials that are not regarded as medical instruments (here: drain - plastic) are not segmented (rule 1). |
| 14 | 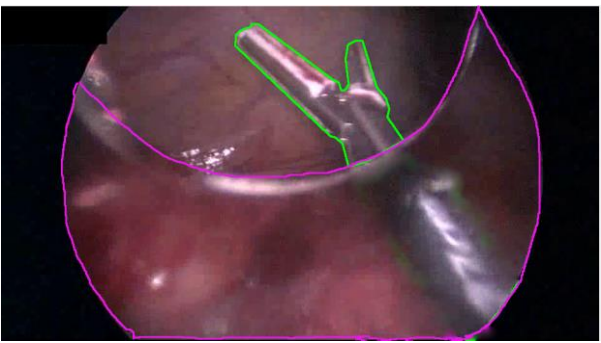 | Tricky example. A medical instrument is visible at the end (opening) of the trocar. The part covered by the transparent trocar is not regarded as a visible part of the instrument (rule 2 an 3). |
| 15 | 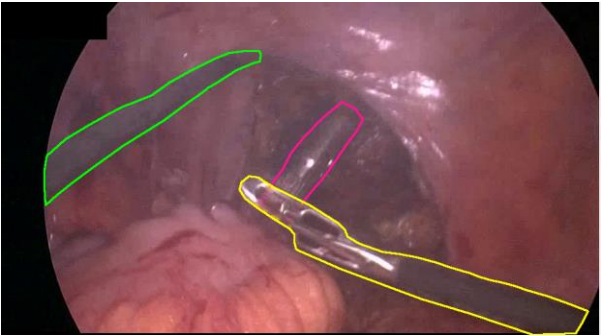 | Even if instruments (pink) are visible through holes of another instrument (yellow), they are not annotated because the holes are regarded as part of the (other) instrument (rule 4). |
| 16 | 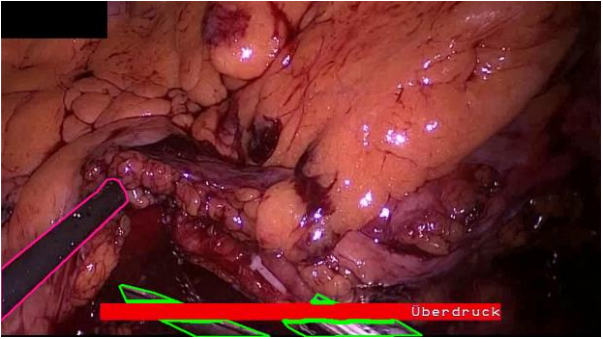 | The instrument represented by the green contour comprises multiple parts due to image overlay (rule 6). |
| 17 | 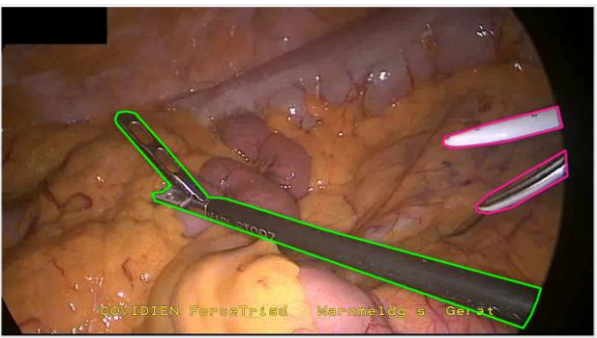 | As text overlays only block small areas of sight and form no matter in line of sight, they are ignored for the segmentation (rule 5). |
| 18 | 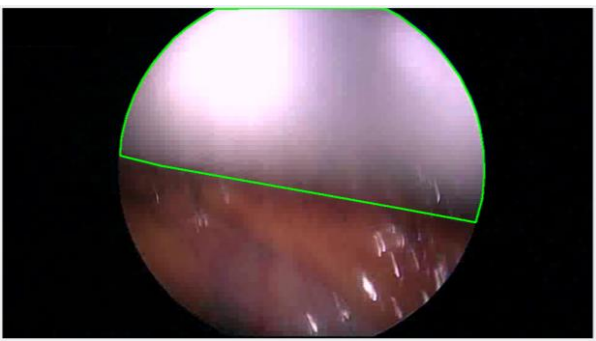 | Instruments are visible from a variety of perspectives. The video sequence helps identify blurred close-ups. |
| 19 | 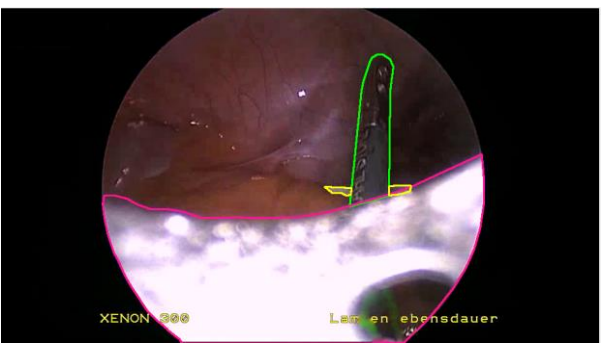 | Even if instruments (green) are visible through holes of another instrument (pink), they are not annotated because holes are regarded as part of the (other) instrument (rule 4). |

| 20 | 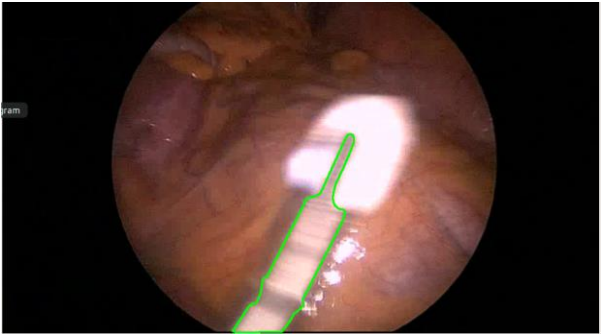 | Motion artefacts pose a challenge to accurate and precise annotation. In this example, the inner part of the blurred area has been identified as instrument. |

## Supplementary Tables

**Supplementary Table 1:** Example of Device Data

| Endoscope | Thermo-flator | | | | | | | OR lights | | | Endoscopic light source | Endoscope | | |
|---|---|---|---|---|---|---|---|---|---|---|---|---|---|---|
| Frame # | Current gas flowrate | Target gas flow rate | Current gas pressure | Target gas pressure | Used gas volume | Gas supply pressure | Device on? | All lights off? | Intensity light 1 | Intensity light 2 | Intensity | White balance | Gains | Exposure index |
| 0 | 115 | 160 | 9 | 15 | 42 | 670 | 0 | 0 | 100 | 100 | 5 | 0 | -1 | -1 |
| 1 | 115 | 160 | 9 | 15 | 42 | 670 | 0 | 0 | 100 | 100 | 5 | 0 | -1 | -1 |
| 2 | 115 | 160 | 9 | 15 | 42 | 670 | 0 | 0 | 100 | 100 | 5 | 0 | -1 | -1 |
| 3 | 115 | 160 | 9 | 15 | 42 | 670 | 0 | 0 | 100 | 100 | 5 | 0 | -1 | -1 |
| 4 | 115 | 160 | 9 | 15 | 42 | 660 | 0 | 0 | 100 | 100 | 5 | 0 | -1 | -1 |
| 5 | 115 | 160 | 9 | 15 | 43 | 660 | 0 | 0 | 100 | 100 | 5 | 0 | -1 | -1 |
| 6 | 115 | 160 | 9 | 15 | 43 | 660 | 0 | 0 | 100 | 100 | 5 | 0 | -1 | -1 |
| 7 | 115 | 160 | 9 | 15 | 43 | 660 | 0 | 0 | 100 | 100 | 5 | 0 | -1 | -1 |
| 8 | 115 | 160 | 9 | 15 | 43 | 650 | 0 | 0 | 100 | 100 | 5 | 0 | -1 | -1 |
| 9 | 115 | 160 | 9 | 15 | 43 | 650 | 0 | 0 | 100 | 100 | 5 | 0 | -1 | -1 |
| 10 | 115 | 160 | 9 | 15 | 43 | 650 | 0 | 0 | 100 | 100 | 5 | 0 | -1 | -1 |
| 11 | 115 | 160 | 9 | 15 | 43 | 650 | 0 | 0 | 100 | 100 | 5 | 0 | -1 | -1 |
| 12 | 115 | 160 | 9 | 15 | 43 | 650 | 0 | 0 | 100 | 100 | 5 | 0 | -1 | -1 |
| 13 | 115 | 160 | 9 | 15 | 43 | 650 | 0 | 0 | 100 | 100 | 5 | 0 | -1 | -1 |
| 14 | 115 | 160 | 9 | 15 | 43 | 650 | 0 | 0 | 100 | 100 | 5 | 0 | -1 | -1 |
| 15 | 115 | 160 | 9 | 15 | 44 | 650 | 0 | 0 | 100 | 100 | 5 | 0 | -1 | -1 |
| 16 | 115 | 160 | 9 | 15 | 44 | 650 | 0 | 0 | 100 | 100 | 5 | 0 | -1 | -1 |
| 17 | 115 | 160 | 9 | 15 | 44 | 650 | 0 | 0 | 100 | 100 | 5 | 0 | -1 | -1 |
| 18 | 115 | 160 | 9 | 15 | 44 | 650 | 0 | 0 | 100 | 100 | 5 | 0 | -1 | -1 |
| 19 | 115 | 160 | 9 | 15 | 44 | 650 | 0 | 0 | 100 | 100 | 5 | 0 | -1 | -1 |
| 20 | 115 | 160 | 9 | 15 | 44 | 650 | 0 | 0 | 100 | 100 | 5 | 0 | -1 | -1 |
| 21 | 115 | 160 | 9 | 15 | 44 | 650 | 0 | 0 | 100 | 100 | 5 | 0 | -1 | -1 |
| ... | ... | ... | ... | ... | ... | ... | ... | ... | ... | ... | ... | ... | ... | ... |
| 371583 | 30 | 190 | 13 | 19 | 137 | 650 | 0 | 0 | 100 | 100 | 70 | 0 | -1 | -1 |
| 371584 | 30 | 190 | 13 | 19 | 137 | 650 | 0 | 0 | 100 | 100 | 70 | 0 | -1 | -1 |
| 371585 | 30 | 190 | 13 | 19 | 137 | 650 | 0 | 0 | 100 | 100 | 70 | 0 | -1 | -1 |
| 371586 | 30 | 190 | 13 | 19 | 137 | 650 | 0 | 0 | 100 | 100 | 70 | 0 | -1 | -1 |
| 371587 | 30 | 190 | 13 | 19 | 137 | 650 | 0 | 0 | 100 | 100 | 70 | 0 | -1 | -1 |
| 371588 | 30 | 190 | 13 | 19 | 138 | 650 | 0 | 0 | 100 | 100 | 70 | 0 | -1 | -1 |
| 371589 | 30 | 190 | 13 | 19 | 138 | 660 | 0 | 0 | 100 | 100 | 70 | 0 | -1 | -1 |
| 371590 | 30 | 190 | 13 | 19 | 138 | 660 | 0 | 0 | 100 | 100 | 70 | 0 | -1 | -1 |
| 371591 | 30 | 190 | 13 | 19 | 138 | 660 | 0 | 0 | 100 | 100 | 70 | 0 | -1 | -1 |
| 371592 | 30 | 190 | 13 | 19 | 138 | 660 | 0 | 0 | 100 | 100 | 70 | 0 | -1 | -1 |
| 371593 | 30 | 190 | 13 | 19 | 138 | 660 | 0 | 0 | 100 | 100 | 70 | 0 | -1 | -1 |
| 371594 | 30 | 190 | 13 | 19 | 138 | 660 | 0 | 0 | 100 | 100 | 70 | 0 | -1 | -1 |
| 371595 | 30 | 190 | 13 | 19 | 138 | 660 | 0 | 0 | 100 | 100 | 70 | 0 | -1 | -1 |
| 371596 | 70 | 190 | 15 | 19 | 138 | 670 | 0 | 0 | 100 | 100 | 70 | 0 | -1 | -1 |
| 371597 | 70 | 190 | 15 | 19 | 138 | 670 | 0 | 0 | 100 | 100 | 70 | 0 | -1 | -1 |
| 371598 | 70 | 190 | 15 | 19 | 138 | 670 | 0 | 0 | 100 | 100 | 70 | 0 | -1 | -1 |
| 371599 | 70 | 190 | 15 | 19 | 138 | 670 | 0 | 0 | 100 | 100 | 70 | 0 | -1 | -1 |
| 371600 | 70 | 190 | 15 | 19 | 138 | 650 | 0 | 0 | 100 | 100 | 70 | 0 | -1 | -1 |
| 371601 | 70 | 190 | 15 | 19 | 138 | 650 | 0 | 0 | 100 | 100 | 70 | 0 | -1 | -1 |
| 371602 | 70 | 190 | 15 | 19 | 138 | 650 | 0 | 0 | 100 | 100 | 70 | 0 | -1 | -1 |

**Supplementary Table 2:** Example of Phase Annotations

| Frame # | Phase ID |
|---|---|
| 0 | 0 |
| 1 | 0 |
| 2 | 0 |
| ... | ... |
| 11838 | 7 |
| ... | ... |
| 17141 | 3 |
| ... | ... |
| 37025 | 7 |
| ... | ... |
| 48726 | 6 |
| ... | ... |
| 73043 | 4 |
| ... | ... |
| 82901 | 5 |
| ... | ... |
| 85844 | 6 |
| ... | ... |
| 102425 | 4 |
| ... | ... |
| 118159 | 8 |
| ... | ... |
| 124054 | 1 |
| ... | ... |
| 133084 | 8 |
| ... | ... |
| 147349 | 4 |
| ... | ... |
| 152145 | 1 |
| ... | ... |
| 156043 | 4 |
| ... | ... |
| 159385 | 8 |
| ... | ... |
| 169154 | 0 |
| ... | ... |
| 172143 | 8 |
| ... | ... |
| 232589 | 9 |
| ... | ... |
| 310724 | 10 |
| ... | ... |
| 340328 | 11 |
| ... | ... |
| 369387 | 12 |
| ... | ... |
| 371602 | 12 |